%% file: root.tex
\documentclass[conference]{IEEEtran}
\IEEEoverridecommandlockouts
\usepackage{graphics} 
\usepackage{graphicx,import, multicol}
\usepackage{subfigure}
\usepackage{algorithmic}
\usepackage{amsmath,amssymb,amsfonts}
\usepackage{cite}

\usepackage{tikz}                      
\usepackage{tikzscale}
\usepackage{relsize}
\usetikzlibrary{calc, backgrounds, shapes, arrows, shapes}
\usepackage{pgfplots}                  
\pgfplotsset{compat=newest} 
\usepgfplotslibrary{external}
\usetikzlibrary{plotmarks}
\def\BibTeX{{\rm B\kern-.05em{\sc i\kern-.025em b}\kern-.08em
		T\kern-.1667em\lower.7ex\hbox{E}\kern-.125emX}}
	
\input{doc/commands}

\title{Extended Existence Probability Using Digital Maps for Object Verification%
}

\author{\IEEEauthorblockN{Fabian Gies, Joachim Posselt, Michael Buchholz and Klaus Dietmayer}
	\IEEEauthorblockA{\textit{Institute of Measurement, Control and Microtechnology} \\%
		\textit{Ulm University}\\%
		89081 Ulm, Germany \\%
		Email: \{fabian.gies, joachim.posselt, michael.buchholz, klaus.dietmayer\}@uni-ulm.de}
	}

\begin{document}

\maketitle
\tikzexternaldisable
\copyrightnotice
\tikzexternalenable
\thispagestyle{empty}
\pagestyle{empty}

\begin{abstract}
    A main task for automated vehicles is an accurate and robust environment perception. Especially, an error-free detection and modeling of other traffic participants is of great importance to drive safely in any situation. For this purpose, multi-object tracking algorithms, based on object detections from raw sensor measurements, are commonly used. However, false object hypotheses can occur due to a high density of different traffic participants in complex, arbitrary scenarios. For this reason, the presented approach introduces a probabilistic model to verify the existence of a tracked object. Therefore, an object verification module is introduced, where the influences of multiple digital map elements on a track's existence are evaluated. Finally, a probabilistic model fuses the various influences and estimates an extended existence probability for every track. In addition, a Bayes Net is implemented as directed graphical model to highlight this work's expandability. The presented approach, reduces the number of false positives, while retaining true positives. Real world data is used to evaluate and to highlight the benefits of the presented approach, especially in urban scenarios.
\end{abstract}

\input{doc/introduction}
\input{doc/system_overview}
\input{doc/verification}
\input{doc/evaluation}
\input{doc/conclusion}
\input{doc/acknowledgment}

\bibliographystyle{IEEEtranBST2/IEEEtran}
\bibliography{IEEEtranBST2/IEEEabrv,bib/2020_object_verification}

\end{document}

%% file: doc/commands.tex
\renewcommand{\vec}[1]{{\mathrm{#1}}}
\newcommand{\vecset}[1]{\boldsymbol{\vec{#1}}}
\newcommand{\vecest}[1]{\hat{\vec{#1}}}
\newcommand{\trans}{^\text{\rm \textup{T}}}

\renewcommand{\figurename}{Fig.~}

\graphicspath{{img/}}

\newlength\figH
\newlength\figW
\newcommand\copyrighttext{%
	\footnotesize \copyright~2020 IEEE. Personal use of this material is permitted. Permission from IEEE must be obtained for all other uses, in any current or future media, including reprinting/republishing this material for advertising or promotional purposes, creating new collective works, for resale or redistribution to servers or lists, or reuse of any copyrighted component of this work in other works.}
\newcommand\copyrightnotice{%
	\begin{tikzpicture}[remember picture,overlay]
	\node[anchor=south,yshift=10pt] at (current page.south) {\fbox{\parbox{\dimexpr\textwidth-\fboxsep-\fboxrule\relax}{\copyrighttext}}};
	\end{tikzpicture}%
}

%% file: doc/introduction.tex
\section{Introduction}
\label{sec:introduction}
For automated vehicles, a complete, robust and accurate perception of the local environment is required. In particular, the detection and modeling of all traffic participants is necessary to enable automated driving. Therefore, the vehicles are equipped with a large variety of different sensors to receive a full and precise depiction of the surrounding. To make use of this huge amount of data and to be able to model other traffic participants, multiple subsequent algorithmic steps are necessary. Commonly, the first step is to create object hypotheses from the raw sensor data for each traffic participant detected, as \cite{Herzog2019, Danzer2019, Arnold2019} show. The object hypotheses are then used as measurements by any multi-object tracking filter \cite{Vo2015} to estimate a precise state. In this work, the Labeled Multi-Bernoulli Filter (LMB) \cite{Reuter2014} estimates a multi-object state based on noisy measurements, while considering clutter and missed detections over subsequent time steps. Besides the spatial state distribution, an existence probability for each track is calculated. Any track exceeding a minimum required existence probability is considered in subsequent modules.

Although these algorithms are well known and have great success, there are many arbitrary scenarios with high densities of different traffic participants where the object detection and consequently the multi-object tracking can fail. As a result, false objects hypotheses are detected or existing objects are absent. Consequently, these false positives or missing objects can lead to wrong assumptions and to unknown behavior during the later processing steps, e.g. a trajectory generation of automated vehicles.

In this work, an extended existence probability is developed with the objective to represent the presence of an object regarding contextual information from digital maps. Due to the additional information, a reduction of falsely detected objects is achieved, while correctly detected objects are retained. In the literature, several ways to integrate map information and to improve an object perception system exist. Hosseinyalamdary et al. \cite{Hosseinyalamdary2015}, proposes an approach, where the point cloud of a LiDAR scan is directly filtered regarding OpenStreetMap (OSM) \cite{OSM2017}. Consequently, the preprocessing of raw sensor data is sped up and erroneous detections are reduced. However, this concept loses information within the first processing steps that cannot be retrieved later on. Hence, map elements that are not depicted in the digital map can result in missing objects. Another approach is presented by Danzer \cite{Danzer2018}, where road information influences the prediction step of a multi-object tracking. The results show, that the estimation accuracy benefits from using contextual information. Since, \cite{Danzer2018} focuses on a vehicle tracking system, this paper proposes an approach to integrate digital map information which is independent regarding the object class. This work's main idea derives from the authors' previous publication \cite{Gies2018}, where a high-level fusion module subsequent to the multi-object tracking is introduced. Besides digital maps, object hypotheses from a dynamic occupancy grid map are considered. The results show reduced missing and false objects due to information fusion. 

This paper focuses on an object verification solely based on contextual information from digital maps, which is integrated into the environment model module. A major reason for the presented system layout is the modularity and scalability, which is developed under the consideration of different approaches introduced by \cite{Nuss2014, Ulbrich2017, Tas2017} for automated vehicles. Integrating digital map information can be challenging in edge cases, hence this work implements probabilistic models to consider uncertainties of the digital map and the tracks' state. Therefore, digital map elements from highly precise mapped roads by Atlatec GmbH \cite{Atlatec2016} and building outlines from OSM \cite{OSM2017} are incorporated. The contextual information are modeled as probabilistic influences and evaluated using an independent influence model (IIM) and a Bayes Net (BN) \cite{Murphy2012}. The models estimate an extended existence probability. Finally, a threshold defines the minimum required extended existence probability and decides whether an object exists or not.

This paper is structured as follows: In Section \ref{sec:system_overview}, an overview of the functional system architecture is given. Here, necessary components of the perception framework are described and motivated. Section \ref{sec:verification} describes the probabilistically modeled influences and introduces the extended existence probability. An evaluation of the algorithm based on real world data is then discussed in Section \ref{sec:evaluation}. Finally in Section \ref{sec:conclusion}, the work is summarized and an outlook is given. 

%% file: doc/system_overview.tex
\section{Functional System Overview}
\label{sec:system_overview}
Since this work presents a component in the latter stages of a perception system, a short insight of 
necessary preprocessing functional modules will be given in the following section. Note that, the presented system architecture depicts only a subset of all components for an automated vehicle, e.g. other sensors, free space modeling or behavior planning are missing. 

\begin{figure}[!t]
	\centering
	\resizebox{0.95\columnwidth}{!}{\includegraphics{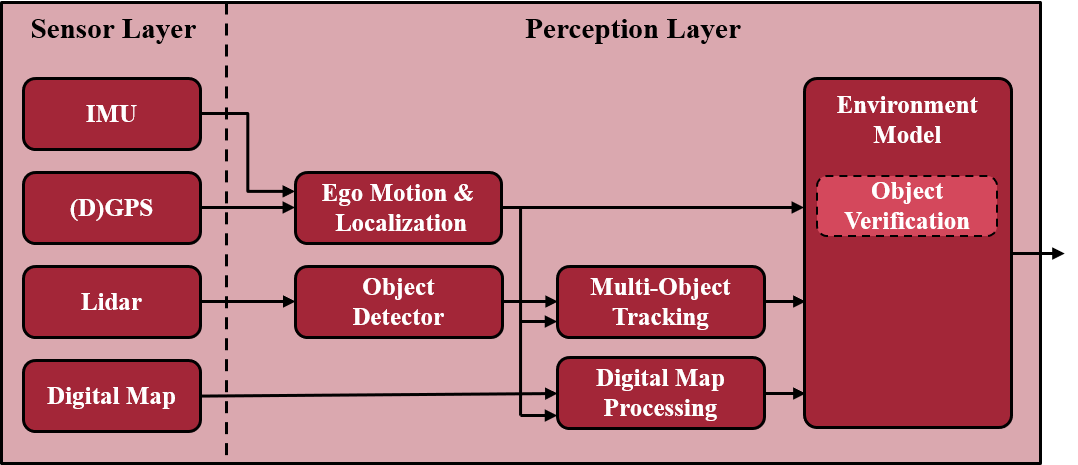}}
	\caption{Implemented functional system architecture of an environment perception introducing the presented object verification module}
	\label{fig:system}
\end{figure}
The introduced object verification module is a part of the perception layer, what is shown in \figurename{\ref{fig:system}}. This system architecture separates the sensor layer and the perception layer. These layers would be followed by an application layer including behavior planning and trajectory generation, but this will not be considered further in this work. In the sensor layer, the sensors are decoded and transmitted to the perception layer. Main characteristics and properties of the perception layer are given in the following. 

\subsection{Ego Motion \& Localization}
First of all, the dynamic states of the automated ego vehicle are essential. Since digital map information is used, the localization of the ego vehicle in a global reference coordinate system is required. Therefore, an Extended Kalman Filter (EKF) filters the measurements of an Inertial Measurement Unit (IMU) and a Differential Global Positioning System (DGPS) and estimates the global ego motion state
\begin{equation}
	\vecest{s}_{\text{ego}} = [x,y,v,a,\varphi,\omega]\trans,
	\label{eq:ego_state}
\end{equation} 
where $x$ and $y$ are the UTM east and north coordinates, $v$~is the absolute velocity and $a$ is the absolute acceleration. Further, $\varphi$ is the UTM orientation and $\omega$ is the yaw rate. In addition, the EKF estimates a full covariance matrix $\vecest{P}_{\text{ego}}$. Inaccuracies of the position and orientation have a direct impact on the coordinate transformation from the vehicle coordinate system to the global coordinate system and vice versa. Due to this, the uncertainty of the localization needs to be considered in a fusion system using map information. Since the ego state \eqref{eq:ego_state} is essential for multi-object tracking and digital map processing, it is directly transmitted. 

\subsection{Digital Map Processing}
\begin{figure}[!t]
	\centering
	\resizebox{0.7\columnwidth}{!}{\includegraphics{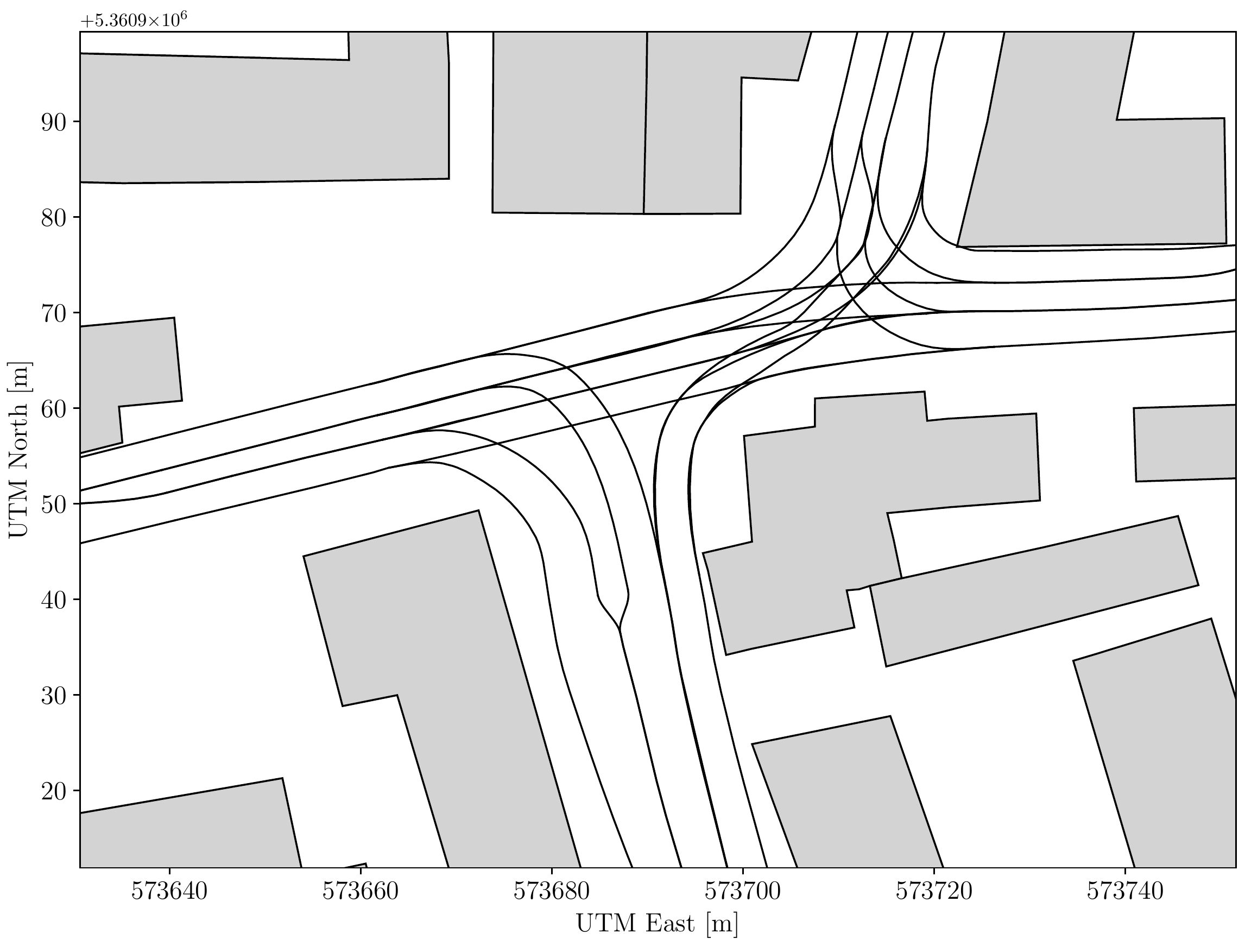}}
	\caption{Sample of the digital map data from the inner city of Ulm. Roads are visualized in black and buildings in grey.}
	\label{fig:digital_map}
\end{figure}
During the digital map processing, a local map section around the ego vehicle is created. Loading and processing the whole map would result in increased latencies and computational costs. In the presented approach, the local map consists of two different map sources. As first source, a highly accurate set of lanes $\vecset{L}$ mapped by Atlatec GmbH \cite{Atlatec2016} are available in the lanelet2 format \cite{Poggenhans2018}. The second source is the OpenStreetMap (OSM) \cite{OSM2017}. OSM data is publicly available and provides a high density of map information, but with unknown inaccuracies and inconsistencies. In the presented approach, the digital map processing module extracts building outlines $\vecset{B}$ that are used to verify the tracked object state. Because these building outlines can overlap in the OSM data, an additional processing step merges overlapping outlines. Finally, the local building outlines are represented as polygons in UTM coordinates. The digital map is shown in \figurename{\ref{fig:digital_map}}, where the roads $\vecset{L}$ (black) and buildings $\vecset{B}$ (grey) are visualized at an intersection in the city center of Ulm, Germany.

\subsection{Object Detector \& Multi-Object Tracking}
\label{seq:tracking}
The multi-object tracking module is supposed to estimate the state of any other road user. Because of temporal filtering, clutter is suppressed and the estimated state is highly accurate. This work proposes the usage of an object detector in the preprocessing, which generates object hypotheses from raw sensor measurements. In the literature, there are a high variety of object detectors for any sensor type \cite{Arnold2019}. Since, this work focuses on the post processing steps of the object verification module, a single object detector is used. Here, the fast object detector for LiDAR point clouds of Herzog \cite{Herzog2019} is implemented. These detections are used as point object measurements. 

In the perception layer, there is no restriction, which multi-object tracking algorithm is used \cite{Vo2015}. Here, an LMB filter of Reuter et al. \cite{Reuter2014} is implemented, which tracks the object detections using a Constant Turn Rate and Acceleration (CTRA) motion model. The filter holds a set $\vecest{S}_{\tau} = \{\vecest{s}_1, ..., \vecest{s}_n \}$ of tracked objects, where
\begin{equation}
	\vecest{s}_{\tau} = [x,y,v,a, \varphi, \omega ]\trans
	\label{eq:track_state}
\end{equation} 
is a single target state vector. The track's state includes a two dimensional position $[x,y]\trans$ and an orientation $\varphi$ at the geometric center point in the ego vehicle coordinate system. Moreover, the absolute velocity $v$, acceleration $a$ and yaw rate $\omega$ are estimated. The covariance matrix~$\vecest{P}_{\tau}$ contains the corresponding variances and covariances of every state. Besides their dynamic states, every tracked object has an unique label~$\ell$, a classification probability~$P_{\vecset{\tau}}(c)$ and an existence probability~$r$. For detailed information on how this existence probability is estimated, refer to \cite{Reuter2014}. During the evaluation, tracks with $r$ exceeding a minimum required threshold $\theta_{r}$, serve as baseline for comparison. Finally, every estimated track is transmitted to the object verification module.

\subsection{Environment Model}
The object verification module is part of the environment model. Here, the main task is to combine multiple information sources and generate a complete and accurate list of objects in the local environment. For the benefit of modular expandability, the algorithm is not integrated into the multi-object tracking. The presented object verification receives data from the ego motion and localization, the digital map processing and the multi-object tracking. As output, the set $\vecest{S}'_{\tau} = \{\vecest{s}_1, ..., \vecest{s}_m \}$ containing every validated track is computed. The single target state vectors~$ \vecest{s}_{\tau} $ are equal compared to the tracks' states from the multi-object tracking. 
However, the number of validated tracks $m$ can be smaller than the number of tracks $n$. In order to decide if an object is valid, an extended existence probability $\eta$ is calculated using different influences from digital map elements. In the end, a final threshold $\theta_{\eta}$ is applied to publish all valid tracks. The subsequent Section \ref{sec:verification} gives a detailed insight on how the extended existence probability is determined.

%% file: doc/verification.tex
\section{Extended Existence Probability for Object Verification}
\label{sec:verification}
The object verification module estimates an extended existence probability for every received track. Therefore, negative and positive influences are modeled probabilistically regarding the track states and digital map elements. In this section, firstly the influences and their calculation are introduced. Secondly, the inference of the extended existence probability incorporating these influences is presented. 

\subsection{Modeling the Digital Map Influences}
In the presented approach, a major design decision is the modeling of multiple influences on the track's existence and evaluating them independently. In this work, only the most effective ones are highlighted. For example, influences modeling, e.g. class depending limited dynamics do not achieve significant improvements and are challenging to parameterize correctly. 
However, influences based on information from the digital map elements $\vecset{B}$ and $\vecset{L}$ have a significant impact. In order to compare a track's state with the digital map, the track's pose \eqref{eq:track_state} is transformed into the UTM coordinate system using the global ego state \eqref{eq:ego_state}. The track's spatial covariance matrix
\begin{equation}
	\vecest{P}'_{\tau} = \left(\begin{array}{rr}
	\sigma_{{\tau},xx}^2 & \sigma_{{\tau},xy}^2 \\
	\sigma_{{\tau},yx}^2 & \sigma_{{\tau},yy}^2
	\end{array} \right),
\end{equation} transforms to a global spatial covariance matrix
\begin{equation}
	\vecest{\Sigma}_{\tau} = R \cdot \vecest{P}'_{\tau} \cdot R\trans + \vecest{P}'_{\text{ego}},
	\label{eq:cov_track_gc}
\end{equation} with the rotation matrix 
\begin{equation}
	R = \left( \begin{array}{rr}
	\cos(\varphi) & \sin(\varphi) \\
	-\sin(\varphi) & \cos(\varphi)
	\end{array} \right).
\end{equation}
Here, $\varphi$ is the UTM orientation from \eqref{eq:ego_state} and $\vecest{P}'_{\text{ego}}$ is the ego motion's spatial covariance matrix. In the following, various influences are described, which consider \eqref{eq:cov_track_gc} and are calculated for every track $\tau$. However, for better readability, the subscript~$\tau$ is omitted in the equations. 

\subsubsection{OSM Building Influence}
\label{sec:osm_influence}
\begin{figure}[!t]
	\centering
	\resizebox{0.7\columnwidth}{!}{\includegraphics{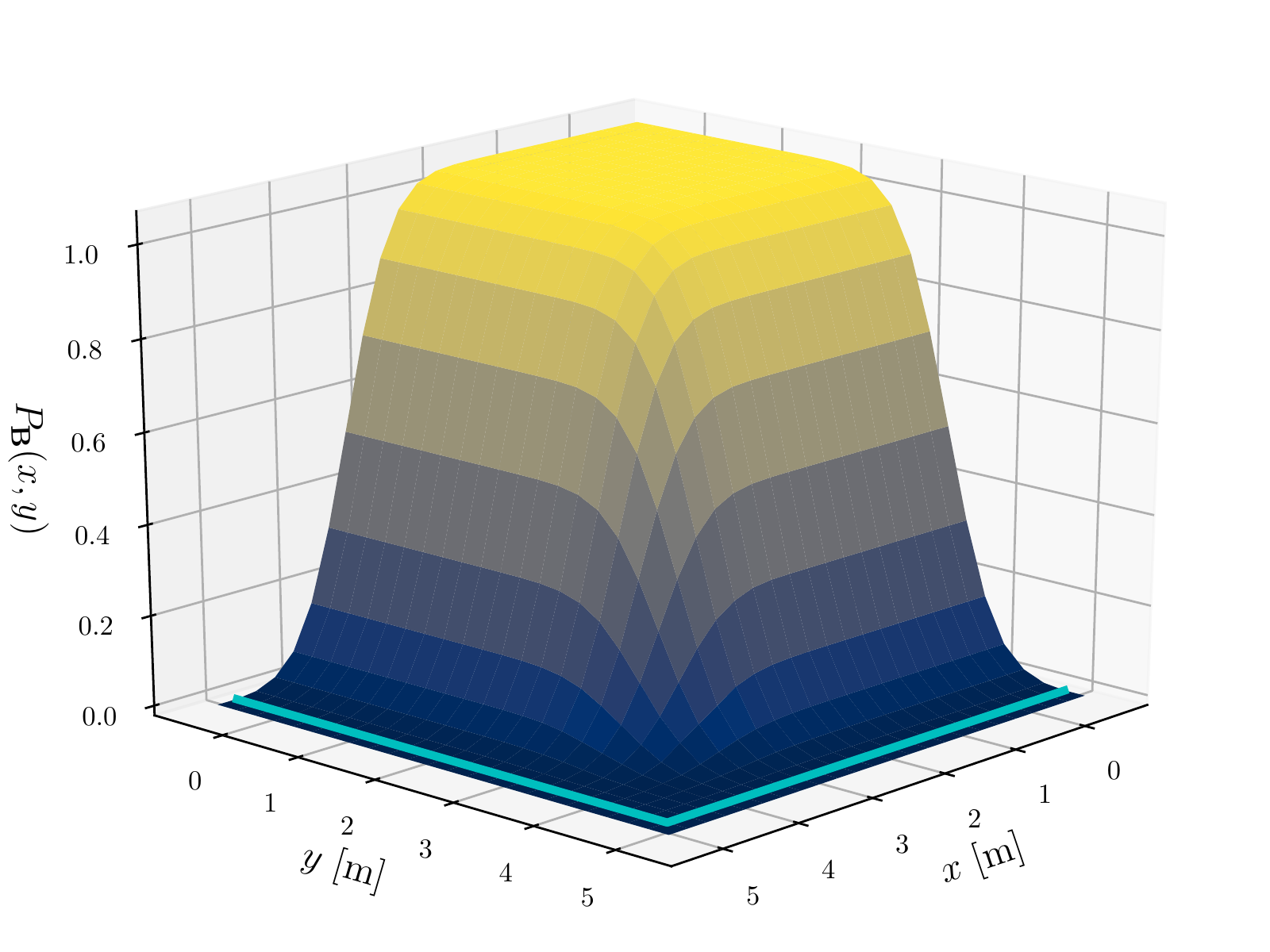}}
	\caption{Building probability function modeling the uncertainty of the map.}
	\label{fig:osm_building}
\end{figure}
A major challenge for camera or LiDAR based sensors are reflections from a building's glass facade, which results in false positive tracks within a building. For that reason, the first influence considers building outlines $\vecset{B}$ to detect a containment and, consequently, negatively influence the extended existence probability. Here, an uncertainty of the OSM mapping process has to be considered. Therefore, similar to Nuss et al. \cite{Nuss2014}, the convolution of the polygons with a multivariate normal distribution $\mathcal{N}(\mu_{\vecset{B}}, \Sigma_{\vecset{B}})$ models the two-dimensional uncertainty of the building outlines. The covariance $\Sigma_{\vecset{B}} = \mathbb{I} \cdot \sigma_{b} $ incorporates a design parameter $\sigma_{b}$, which defines a transition from the building outline to the required depth inside of a building. With the building function 
\begin{equation}
	b(x,y) = 
	\begin{cases}
		1, & \text{when } (x,y) > 3\sigma_{b} \text{ inside building} \\
		0, & \text{otherwise}
	\end{cases},
\end{equation}
the resulting probability function
\begin{equation}
	P_{\vecset{B}}(x,y) = \int\limits_{\mathbb{R}^2}^{} b(x,y) \cdot \mathcal{N}((x,y); (x_0, y_0), \Sigma_{\vecset{B}})
\end{equation}
estimates the probability if a point $(x,y)$ is inside a building. In short, when a point $(x,y)$ is more than $3\sigma_{b}$ within a building's outline, the probability is $P_{\vecset{B}}(x,y) = 1$. This probability function is visualized in \figurename{\ref{fig:osm_building}}. 

\begin{figure}[!t]
	\centering
	\resizebox{0.75\columnwidth}{!}{\includegraphics{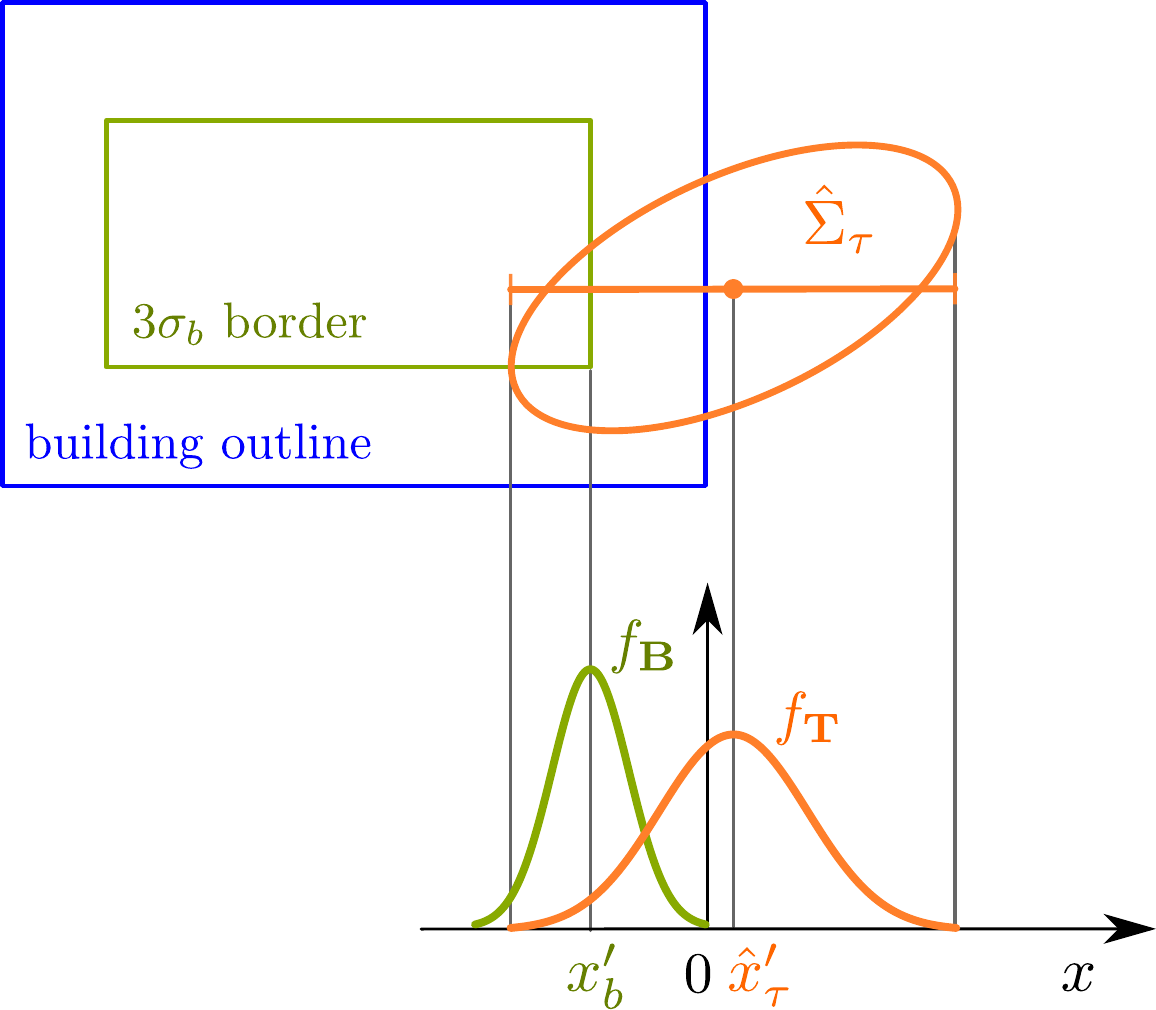}}
	\caption{Building and track uncertainty at reduced dimension along a perpendicular line between track position and building outline.} 
	\label{fig:osm_building_prop}
\end{figure}

Besides uncertainties of the OSM map, the track's global covariance~\eqref{eq:cov_track_gc} has to be considered. Therefore, a perpendicular line between the track's global position and the intersection of the closest building outline is calculated. Afterwards, the two-dimensional problem is reduced to one dimension along this perpendicular line, by reducing the spatial covariance matrix~\eqref{eq:cov_track_gc}. Therefore, the eigen value decomposition and evaluation of the standard ellipsoid function of the spatial covariance along this perpendicular line is calculated. The result is a spatial probability density function (PDF) $f_{\vecset{T}}(x) = \mathcal{N}(x; \hat{x}_{\tau}', \sigma_{\tau}'^2)$ as normal distribution along the perpendicular line, with the track's projected mean $\hat{x}_{\tau}'$ and variance $ \sigma_{\tau}'^2 $. The building outline PDF $f_{\vecset{B}}(x) = \mathcal{N}(x; x_{b}', \sigma_{b}^2)$ is reduced to the same dimension and is defined as normal distribution along this line. The mean is set to $x_{b}' = -3\sigma_{b}$. For clarification, \figurename{\ref{fig:osm_building_prop}} shows an example of the probability functions, where the two-dimensional covariance matrix of the track is reduced to the perpendicular line. The dimension reduction along the perpendicular line results in an approximation error, and due to that, an overestimation of the probability at the building's edges can occur. For simplification, this approximation error is neglected.
Finally, the building containment probability 
\begin{equation}
	\label{eq:cont_prob}
	P_{\vecset{C}}(x) = P_{\vecset{C}}(\hat{x}_{\tau}' \leq x_{b}') = \int\limits_{-\infty}^{\infty}  F_{\vecset{T}}(x) \cdot f_{\vecset{B}}(x) \; dx,
\end{equation}
is the integral over the building's PDF $f_{\vecset{B}}(x)$ and the cumulative distribution function 
\begin{equation}
	\label{eq:cdf}
	F_{\vecset{T}}(x_0) = \int\limits_{-\infty}^{x_0} f_{\vecset{T}}(x) dx'
\end{equation}
of the track's PDF $f_{\vecset{T}}(x)$. The containment probability defines the probability of the track's position $\hat{x}_{\tau}'$ being smaller than the building outline position $x_{b}'$. 
In the presented approach, \eqref{eq:cont_prob} has a decreasing impact on the tracks extended existence probability and, therefore, defines a negative influence.

\subsubsection{Lanelet Influence}
\label{sec:ll_influence}
The second source of digital map information is the lanelet map with lanes $\vecset{L}$. Traffic participants that are near streets, are in most cases relevant objects and, consequently, the lanes define a positive influence and increase the extended existence probability depending on the track's pose relative to the lane. These influences are designed to confirm true tracks. The modeling of the lanes $\vecset{L}$ is split into four sublevels. Thus, tracks are evaluated if they are on the road or near the road and if they are correctly positioned or aligned relative to the lane. Here, a road refers to the entity of all parallel lanes. Evaluating the lanelet influence is similar to the building influence. First, a perpendicular line between the nearest lane border and the track is calculated and, subsequently, the track's spatial covariance matrix \eqref{eq:cov_track_gc} is reduced along this dimension to define the PDF $f_{\vecset{T}}(x) =~\mathcal{N}(x; \hat{x}_{\tau}', \sigma_{\tau}'^2)$. 

Starting with the evaluation if a track is on the road, the road's width $w_r$ is considered. Using the cumulative distribution function \eqref{eq:cdf}, the probability 
\begin{equation}
	P_{\vecset{OR}}(x) = F_{\vecset{T}}(0) - F_{\vecset{T}}(-w_r)
\end{equation}
indicates, whether a track is on the road or not.

Secondly, tracks near the road, e.g. pedestrians on the sidewalk, should be validated and positively weighted. Therefore, a design parameter $\sigma_{r}$ defines a transition width of the road boundary and the normal distribution $f_{\vecset{R}}(x) =~\mathcal{N}(x; x_{r}', \sigma_{r}^2)$ models the road boundary similar to the building outline. As a result, the probability 
\begin{equation}
 	P_{\vecset{NR}}(x) = P_{\vecset{NR}}(\hat{x}_{\tau}' \leq x_{r}') =  \int\limits_{-\infty}^{\infty}  F_{\vecset{T}}(x) \cdot f_{\vecset{R}}(x) \; dx
\end{equation}
indicates if a track is near the road and should be validated by using the cumulative distribution function \eqref{eq:cdf}.

\begin{figure}[!t]
	\centering
	\subfigure[\label{subfig:a:on_road}]{\resizebox{0.45\columnwidth}{!}{\includegraphics{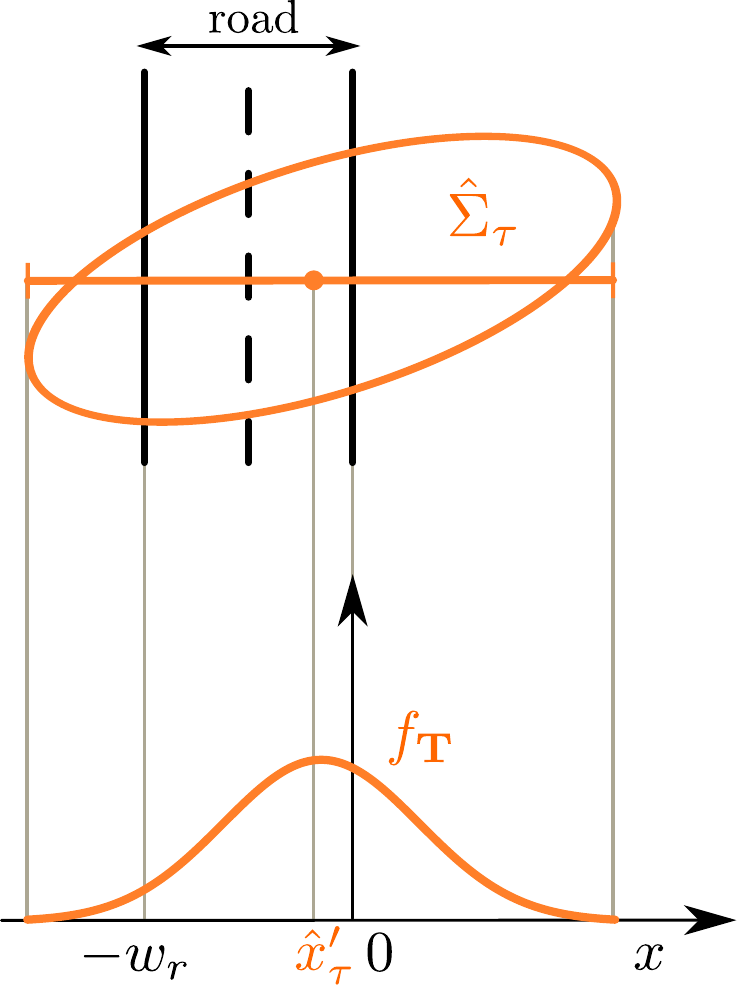}}}
	\subfigure[\label{subfig:a:near_road}]{\resizebox{0.45\columnwidth}{!}{\includegraphics{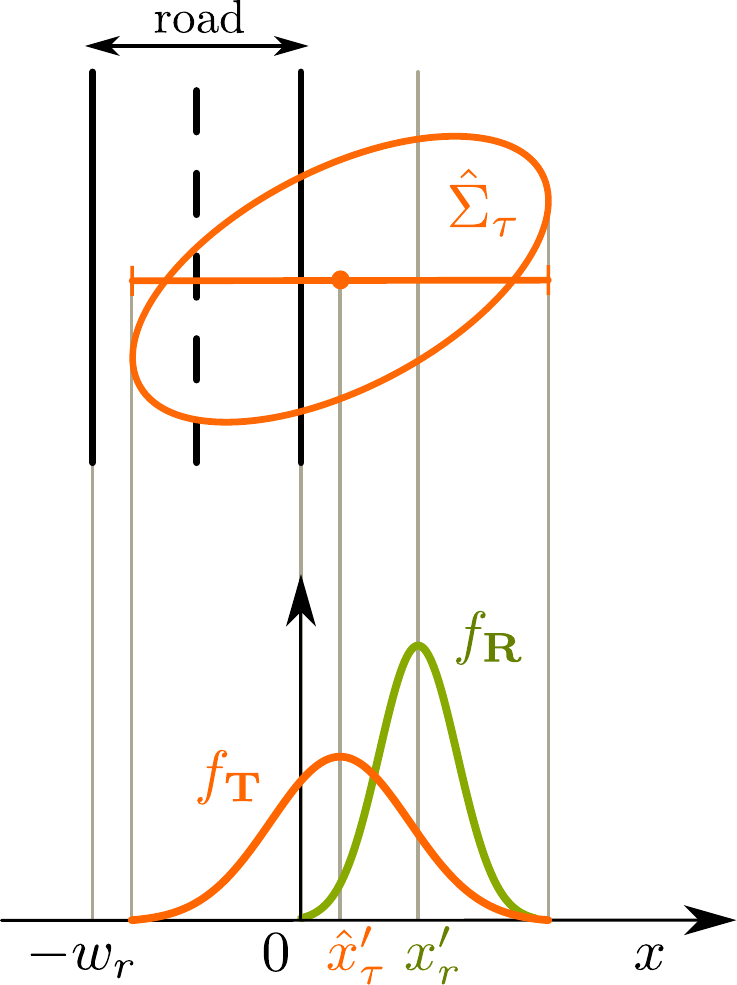}}}
	\caption{On road (a) and near road (b) modeled uncertainties on a perpendicular line between road border and track position.}
	\label{fig:road_probs}
\end{figure}

The third sub level is designed to positively influence vehicles that are located within the boundaries of their associated lane. Therefore, a perpendicular line within the lane boundary is calculated and a normal distribution $f_{\vecset{L}}(x) =~\mathcal{N}(x; x_{l}', \sigma_{l,x}^2)$ across the lane is assumed. Here, the mean $x_{l}' = -\frac{w_l}{2}$ is set to the lane center point and the variance  $\sigma_{l,x}^2 = (\frac{w_l}{6})^2$ is defined such that $3\sigma_{l,x}$ lies on the boundary. This results in the probability 
\begin{equation}
	\label{eq:lane_pos_prob}
	P_{\vecset{LP}}(x) = \int\limits_{-\infty}^{\infty}  f_{\vecset{T}}(x) \cdot f_{\vecset{L}}(x) \; dx,
\end{equation}
which models the tracks' positioning related to the associated lane center point. 

The last probability based on the lanelet map, evaluated the track's orientation related to the course of its associated lane. In consequence, tracks following the lane course are positively influenced. Here, the normal distribution $f_{\vecset{L}}(\Delta\varphi) =~\mathcal{N}(\Delta\varphi; 0, \sigma_{l, \varphi}^2)$ models the distribution over the orientation difference $\Delta\varphi = \hat{\varphi}_{\tau}' - \varphi_{l}'$, where $\varphi_{l}'$ is the lane's course and $\varphi_{\tau}'$ the track's orientation. The mean is set to zero and the variance is $\sigma_{l, \varphi}^2 = (\frac{\pi}{6})^2$. As a result, the orientation difference evaluates between $-\frac{\pi}{2}$ and $\frac{\pi}{2}$. 
Furthermore, with the tracks orientation distribution $f_{\vecset{T}}(\Delta\varphi) =~\mathcal{N}(\Delta\varphi; \Delta\varphi, \sigma_{\tau, \varphi\varphi}^2)$, the lane alignment probability
\begin{equation}
	\label{eq:lane_align_prob}
	P_{\vecset{LA}}(\Delta\varphi) = \int\limits_{-\infty}^{\infty}  f_{\vecset{T}}(\Delta\varphi) \cdot f_{\vecset{L}}(\Delta\varphi) \; d\Delta\varphi,
\end{equation}
indicates a similarity between the track's orientation and the associated lane's course. Obviously, overtaking or backwards moving tracks have an orientation difference greater than $|\frac{\pi}{2}|$ and, subsequently, their probability is zero. Since, the probability \eqref{eq:lane_align_prob} only models positive influences these tracks will not be removed and this behavior can be neglected. 

Since a track can be associated to multiple lanes, the probabilities of \eqref{eq:lane_pos_prob} and \eqref{eq:lane_align_prob} are evaluated for every possible lane. In the end, the lane with the highest sum of both probabilities is taken for further processing.

\subsection{Estimating the Extended Existence Probability}
In the previous subsections, five probability models regarding the OSM building $\vecset{B}$ and the lanelet map information $\vecset{L}$ have been introduced. For further processing, the probabilities are separated into positive $i^{+}$ and negative $i^{-}$ influences. Tracks within a building's outline could be false positives, hence the building probability is defined as negative influence $P(i^{-}_{1}) := P_{\vecset{C}}(x)$. In contrast, the lanelet influences are developed to confirm tracks and, as a result, their probabilities have a positive influence on the track's existence, so that $P(i^{+}_{1}) := P_{\vecset{OR}}(x)$, $P(i^{+}_{2})~:=~P_{\vecset{NR}}(x)$, $P(i^{+}_{3}) := P_{\vecset{LP}}(x)$ and $P(i^{+}_{4}) := P_{\vecset{LA}}(\Delta\varphi)$. These positive and negative influences are fused to estimate the extended existence probability~$\eta$. Therefore, two different approaches are proposed.

First, an independent influence model (IIM) is developed. The main idea of the IIM is the estimation of $\eta$ without considering any correlations between any influence. The IIM combines negative influences with the conditional distribution 
\begin{equation}
	P(i^{-}) := P(i^{-} | P(i^{-}_1),...,P(i^{-}_n)) = \prod_{k=1}^{n} 1 - P(i^{-}_k) .
\end{equation}
Since only the building probability proved to be suitable as negative influence, this distribution reduces to
\begin{equation}
	P(i^{-}) := P(i^{-} | P(i^{-}_1) ) := 1 - P_{\vecset{C}}(x).
\end{equation}
On the other hand, the positive influences are modeled as an equally weighted accumulated average 
\begin{equation}
P(i^{+}) := P(i^{+} | P(i^{+}_1),...,P(i^{+}_m)) = \frac{1}{n} \sum_{k=1}^{m} P(i^{+}_k).
\end{equation}
Using the IIM, the extended existence probability 
\begin{equation}
	\eta = \frac{P(i^{-}) + P(i^{+})}{2}
\end{equation}
calculates an average of negative and positive influences. If a track is not effected by any influence, the extended existence probability $\eta = 0.5$. In addition, heuristic weights could parameterize the impact of any influence, but for reducing design parameters, this is not considered. 

\begin{figure}[!t]
	\centering
	\resizebox{0.7\columnwidth}{!}{\includegraphics{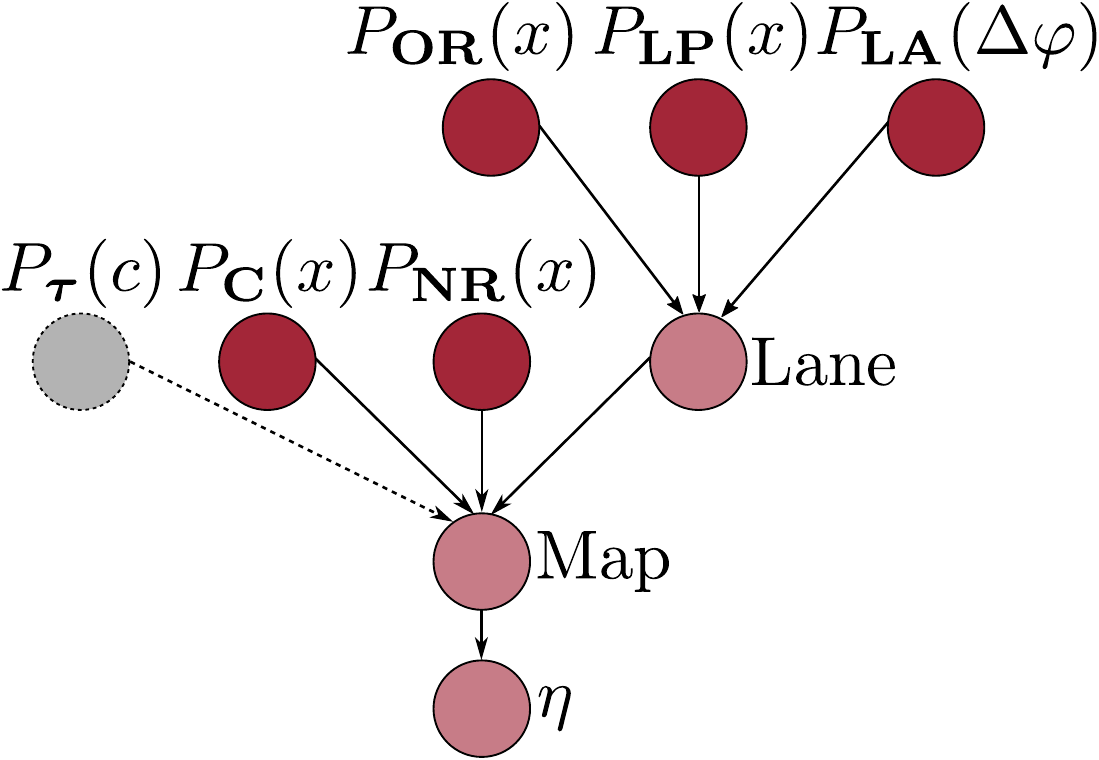}}
	\caption{Graphical model of the Bayes Net with observed nodes (red), hidden nodes (light red) and an optional extension using a classification probability as observed node (grey).}
	\label{fig:bayes_net_graph}
\end{figure}

As second approach, a Bayes Net (BN) \cite{Murphy2012} is implemented. The BN is a well known method for modeling random variables and calculating the joint distribution. Furthermore, the BN generalizes the IIM and is able to consider weights and dependencies between the influences. Therefore, a directed graphical model is designed with analytic expertise by structuring influences and preventing cycles. \figurename{\ref{fig:bayes_net_graph}} visualizes the proposed graphical model. This BN consists of a base graph (red nodes), with all introduced components. The graphical model merges multiple influences into superordinate nodes by combining influences from the lane and map. 
Additionally, \figurename{\ref{fig:bayes_net_graph}} depicts an extension (grey node) using the classification probability $P_{\vecset{\tau}}(c)$ of a track as neutral influence. This extension is intended to illustrate how the BN can easily be scaled up. Finally, after defining all conditional probability tables, the extended existence probability $\eta$ can be inferred. 

In the end, after an inference of the IIM, the base BN or the BN with classification extension (BNe), an corresponding extended existence probability~$\eta$ is appended to every track~$\tau$. A minimum threshold $\theta_{\eta}$ is defined to evaluate if a track exists. If the extended existence probability $\eta < \theta_{\eta}$ is below the threshold, the track will be removed from the set $\vecest{S}_{\tau}$. 

%% file: doc/evaluation.tex
\section{Evaluation}
\label{sec:evaluation}
In this section, an evaluation of the proposed algorithm is given. Therefore, the different approaches IIM, BN and BNe are compared to each other and to a baseline. As baseline, the conventional approach using a threshold $\theta_{r}$ for the existence probability $r$ of the LMB filter is used.

\subsection{System Setup \& Dataset}
The following evaluation uses real world date recorded by the experimental vehicle of Ulm University \cite{Kunz2015}. The ego motion estimation and localization use a highly precise Automotive Dynamic Motion Analyzer (ADMA) and a DGPS. For environment perception, the LiDAR sensor Velodyne VLP-32 is mounted on the vehicle's roof at the front. Object detections generated from the sensor's measurements are tracked with an LMB filter as described in Section \ref{seq:tracking}. Since the KITTI dataset, which is used to train the detector, only provides labels in the front of the vehicle, traffic participants on the sides and back of the vehicle cannot be detected reliably. In consequence, even after LMB filtering, false or missing objects can occur. However, a compensation of the dataset's characteristics during evaluation has no meaningful impact on the effectiveness of the presented approach. The algorithm is implemented within the robot operating system (ROS) framework using C++.

The evaluation includes two different datasets, where the estimated tracks are manually labeled as true positive or false positive. In consequence, undetected objects are not considered. The first dataset scenario takes place in the inner city of Ulm in Germany and consists of 13,065 samples with 8,302 true positives and 4,763 false positives. Each sample represents a track at a single time step in the whole sequence. A major challenge in the urban area are false measurements occurring at glass facades. In consequence, the object detector and LMB filter fail and produce multiple false positives. The second scenario is recorded at a rural suburban area near Ulm and consists of 5,634 samples with 2,936 true positives and 2,698 false positives. Compared to the urban sequence, higher velocities, more vegetation and less buildings are present.

In the presented work, two design parameters are defined in Section \ref{sec:verification}. For the following evaluation, $\sigma_{b} = \frac{1}{3} \; \text{m}$ and $\sigma_{r} = 1.0 \; \text{m}$ led to the best results.

\subsection{Evaluation on Real World Data}
\begin{figure}[!t]
	\centering
	\subfigure[\label{subfig:a:eval_roc_full}]{\resizebox{0.95\columnwidth}{!}{\includegraphics{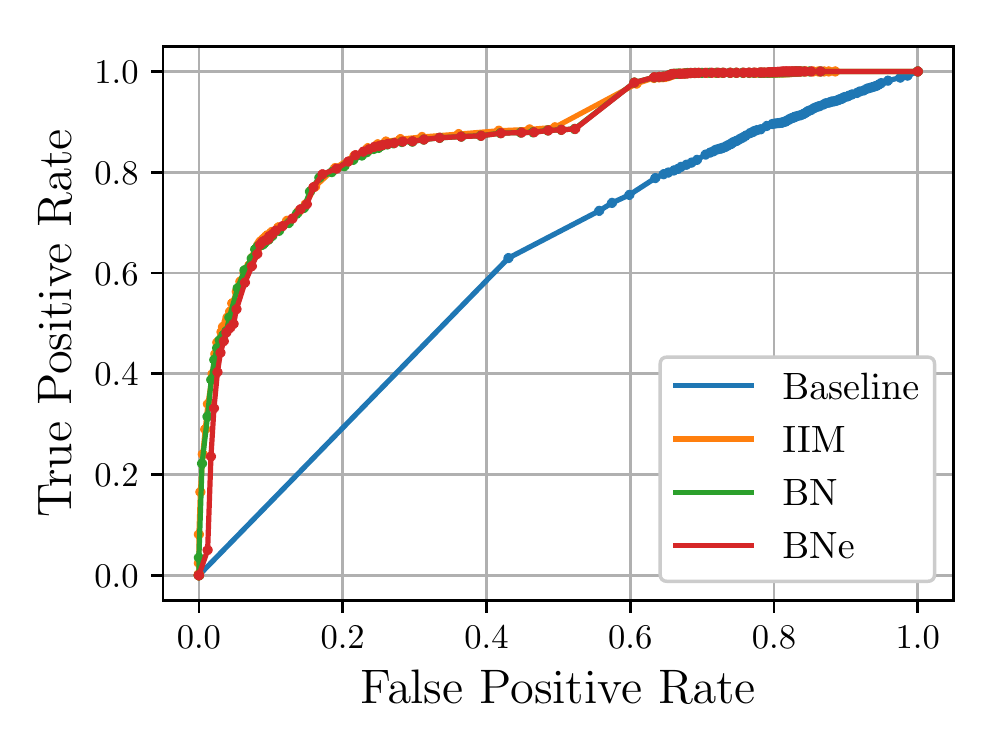}}}
	\subfigure[\label{subfig:a:eval_roc_zoom}]{\resizebox{0.95\columnwidth}{!}{\includegraphics{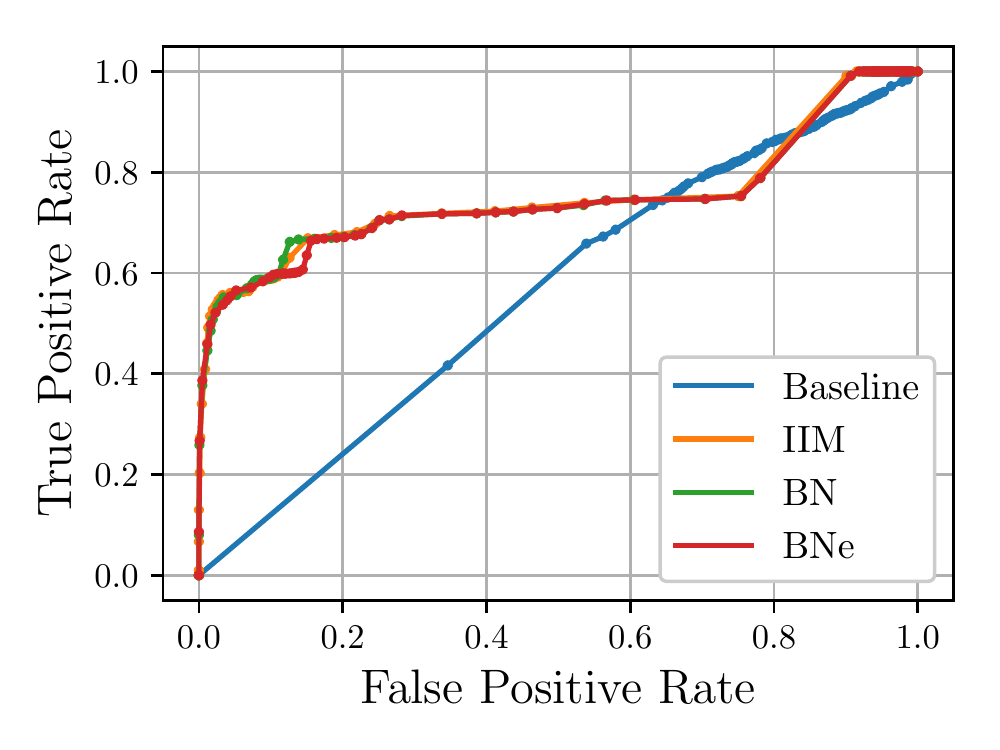}}}
	\caption{ROC for step-wise threshold increment of $\theta_{\eta}$ and $\theta_{r}$ at city scenario (a) and rural scenario (b). The baseline and the three proposed approaches estimating the extended existence probability are shown.}
	\label{fig:eval_roc}
\end{figure}

The main goal of the algorithm is the reduction of false positives while retaining all true positives. Depending on the selected extraction threshold, true positives can falsely be removed and, in consequence, false negatives emerge. On the other hand, when false positives are removed, true negatives emerge, leading to better results with a lower false positive rate. In consequence, a trade-off between falsely removed tracks and correctly removed tracks is required. This behavior is shown and discussed in the following.

As evaluation metrics, the true positive rate and the false positive rate are calculated and a receiver operating characteristics (ROC) is generated. For further details on this common evaluation metric, see Fawcett's work \cite{Fawcett2006}. The ROC is created by a step-wise increment of $0.01$ of the thresholds $\theta_{\eta} \in [0,1]$ used by the IIM, BN and BNe and $\theta_{r} \in [0,1]$ used by the baseline. Starting from Zero, the conditions $\eta \geq \theta_{\eta}$ and $r \geq \theta_{r}$ are always true and in consequence, no tracks are removed. On the other hand, when the thresholds are set to One, the conditions are always false and every track will be removed. The resulting ROC for the city and the rural dataset are shown in \figurename{\ref{fig:eval_roc}}.

The ROC show, that the presented approaches using the IIM, BN or BNe achieve better results than the baseline regarding the false positive rate. Especially in the city scenario, a lot of false tracks within buildings can be removed. Since, in the rural area these buildings are less present the impact is lower. Furthermore, the lanelet influences confirm true tracks near and on a road and consequently, prevent false negatives in these areas. In the rural scenario, a steep increase of the IIM, BN and BNe can be seen, which occurs at an decreasing threshold, where almost every track is verified.  Overall, the IIM, BN or BNe differ only slightly. That is because, the BN is a generic model and inferences the joint distribution under the assumption of conditional independent probabilities similar the IIM. Furthermore, the BNe is a minor extension to the BN regarding a classification, but the dataset mainly contains tracked vehicles and, consequently, this extension has a negligible impact on the results. In summary, these models achieve almost equal results but vary in their modeling, scalability and processing.

The differences between the presented approaches are highlighted by evaluating  the precision, recall and accuracy at certain operation point with the thresholds $\theta_{\eta} = 0.35$ and $\theta_{r} = 0.05$. These thresholds are chosen to maximize the true positive rate, while minimizing the number of false negatives, what corresponds to the top left corner in the ROC. The resulting set of tracks $\vecest{S}'_{\tau}$ would be transmitted to the behavior and trajectory planning and false negatives can lead to arbitrary behavior and possibly fatal consequences. That is why, choosing the threshold $\theta_{\eta}$ should focus on minimizing the number of false negatives. The three approaches IIM, BN and BNe do differ by less than $1\%$ as described above. Compared to the baseline, especially, the recall is approximately $10\%$ higher in both sequences. In consequence of multiple false positives in the city area occurring within buildings, the precision is $6\%$ higher and the accuracy $10\%$. Whereas in the rural area, the precision and accuracy differentiate only around $2\%$.
Finally, the BN can easily be extended with more influences like the BNe. As a consequence, this approach is recommended.


%% file: doc/conclusion.tex
\section{Conclusion}
\label{sec:conclusion}
In summary, the presented approach introduces an algorithm to incorporate digital map information into an extended existence probability of tracked traffic participants. Therefore, multiple probabilistic models define influences of map elements. Furthermore, an independent probabilistic model and a Bayes Net infer the introduced influences to estimate an extended existence probability. Based on this probability, false tracks can be removed, while keeping true tracks and, consequently, the false positive rate is reduced. Finally, an evaluation on real world data in an city and rural area show the significance and performance of the presented approach. 

For future work, the algorithm can be extended with more map elements, e.g. crosswalks or bicycle lanes regarding vulnerable road users. Here, the Bayes Net should be applied and the optimal graphical model can be trained.

%% file: doc/acknowledgment.tex
\section*{Acknowledgment}
This work was performed as part of Joachim Posselts' research within a thesis to obtain the Bachelor of Science. The authors thank the input of his colleague Sebastian Bitzer. 

The research leading to these results was conducted within the Tech Center a-drive. Responsibility for the information and views set out in this publication lies entirely with the authors.

Part of this research was accomplished within the project UNICARagil (FKZ 16EMO0290). We acknowledge the financial support for the project by the Federal Ministry of Education and Research of Germany (BMBF).